\documentclass{article}
\usepackage{amsmath,graphicx}
 \usepackage[preprint]{spconf}
\usepackage{subfig}
\usepackage[export]{adjustbox}
\usepackage{cite}
\usepackage{balance}

\usepackage{tikz}
\usepackage{lipsum}

\copyrightnotice{\begin{tabular}[t]{@{}l@{}}© 2020 IEEE.  Personal use of this material is permitted. Permission from IEEE must be obtained for all other uses, in any current or\\future media, including reprinting/republishing this material for advertising or promotional purposes, creating new collective works, for\\resale or redistribution to servers or lists, or reuse of any copyrighted component of this work in other works.\end{tabular}}

\title{GLEASON GRADING OF HISTOLOGY PROSTATE IMAGES THROUGH SEMANTIC SEGMENTATION VIA RESIDUAL U-NET}

\begin{document}

\name{%
\begin{tabular}{@{}c@{}}
Amartya Kalapahar$^{\star}$,
Julio Silva-Rodr\'iguez$^{\dagger }$,  
Adri\'an Colomer$^{\star}$,
Fernando López-Mir$^{\star}$ and
Valery Naranjo$^{\star}$ 
\thanks{This work was supported by the Spanish Ministry of Economy and Competitiveness through project DPI2016-77869. The work of Fernando López-Mir has been supported by the Polytechnic University of Valencia (Grant PAID-10-18). The Titan V used for this research was donated by the NVIDIA Corporation.}
\end{tabular}}

\address{%
\begin{tabular}{cc}
Institute of Research and Innovation in Bioengineering$^{\star}$ &
Institute of Transport and Territory$^{\dagger}$ \\ 
\textit{Universitat Polit\`ecnica de Val\`encia}, Spain & 
\textit{Universitat Polit\`ecnica de Val\`encia}, Spain \\
vnaranjo@dcom.upv.es &
jjsilva@upv.es
\end{tabular}}

\maketitle

\begin{abstract}
Worldwide, prostate cancer is one of the main cancers affecting men. The final diagnosis of prostate cancer is based on the visual detection of Gleason patterns in prostate biopsy by pathologists. Computer-aided-diagnosis systems allow to delineate and classify the cancerous patterns in the tissue via computer-vision algorithms in order to support the physicians' task. The methodological core of this work is a \textit{U-Net} convolutional neural network for image segmentation modified with residual blocks able to segment cancerous tissue according to the full Gleason system. This model outperforms other well-known architectures, and reaches a pixel-level Cohen's quadratic Kappa of $0.52$, at the level of previous image-level works in the literature, but providing also a detailed localisation of the patterns.  
\end{abstract}


\section{Introduction}
\label{sec:introduction}
\begin{figure}[htb]
    \centering
      \subfloat[\label{fig1a}]{\includegraphics[width=.235\linewidth, frame]{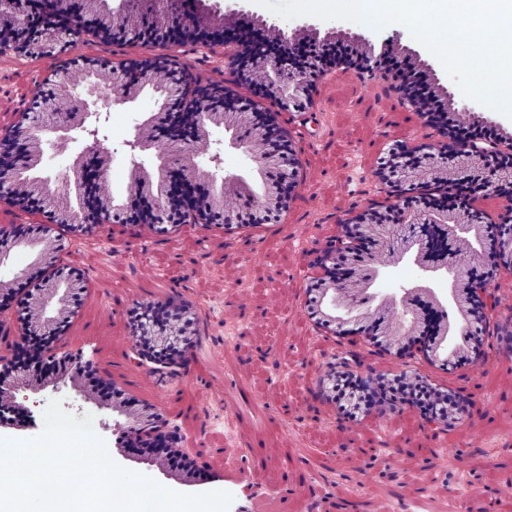}}
      \hspace*{\fill}
      \subfloat[\label{fig1b}]{\includegraphics[width=.235\linewidth, frame]{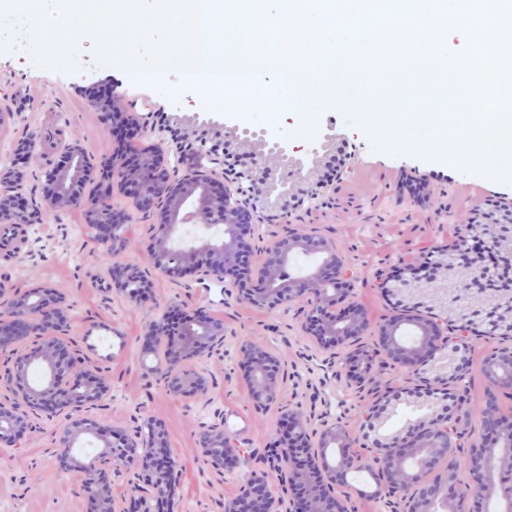}}
      \hspace*{\fill}
      \subfloat[\label{fig1c}]{\includegraphics[width=.235\linewidth, frame]{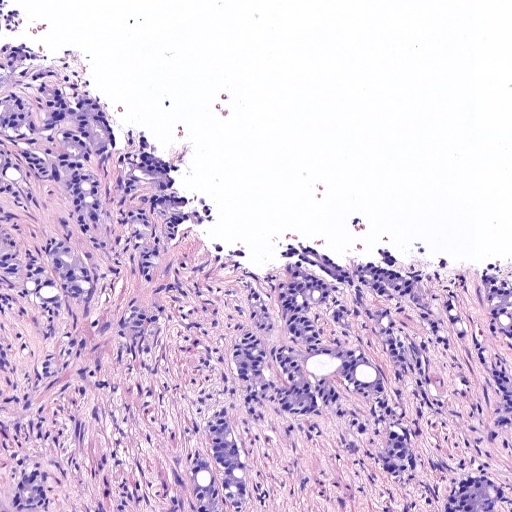}}
      \hspace*{\fill}
      \subfloat[\label{fig1d}]{\includegraphics[width=.235\linewidth, frame]{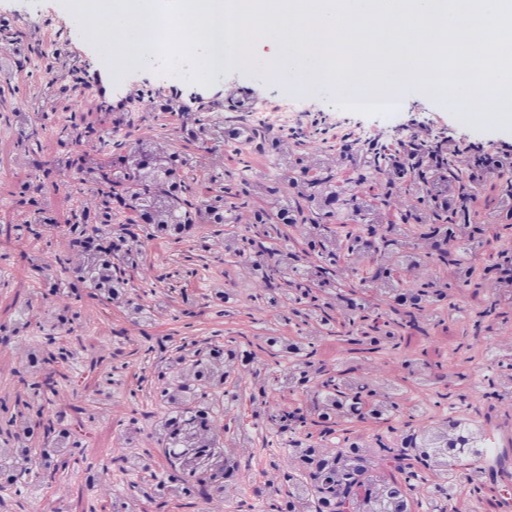}}
    \caption{Examples of histology prostate regions. (a): Non-cancerous glands, (b): Gleason pattern $3$, (c): Gleason pattern $4$ and (d) Gleason pattern $5$.}
    \label{fig1}
\end{figure}

Prostate cancer is the second most common cancer in men \cite{Ferlay2015Cancer2012} and new cases account for $21\%$ of all new cancer diagnoses in men each year \cite{Siegel2016Cancer2016}. The Gleason grading system is widely accepted as a part of a standard protocol when determining the severity of cancer and is related to the growth pattern of tumour glands \cite{gleason}. The system consists of three grades, from $3$ to $5$, each of them including clusters of glandular patterns (or Gleason patterns, referred to in this paper also as $GP$) with similar prognosis (see Fig. \ref{fig1}). In the clinical practice, the stained prostate biopsies are analysed by visual inspection by the pathologists, in order to detect cancerous patterns. Evaluating every single sample manually is a very time-consuming and subjective task \cite{Litjens2016DeepDiagnosis}. For this reason, in recent years, the use of con computer-aided-diagnosis tools based on computer-vision algorithms has experimented a growth in this field.       

Previous works in the literature have carried out different strategies in order to analyse prostate biopsy images. There are three main approaches for prostate histology images processing: image-level predictions \cite{Arvaniti2018AutomatedLearning,Nir2019ComparisonImages} , pixel-level segmentation \cite{Li2019PathImages,Ing2018SemanticNetworks,Li2017AProstatectomies} or gland-level analysis \cite{Garcia2019First-stageLearning}. The image-level processing provides a general cancerous pattern for a region, lacking a precise delineation of the structures in the tissue, while the gland-level analysis is limited to cancerous patterns with glandular structures (i.e. Gleason pattern $3$ or $4$). Contrary, the pixel-level semantic segmentation can work over all the different cancerous patterns, and provides a precise delimitation of the cancerous patterns in the tissue. Nevertheless, one limiting factor in prior works regarding semantic segmentation of Gleason patterns is the low prevalence of Gleason pattern $5$. Despite the fact that image-level studies have been able to full gradation of Gleason patterns \cite{Arvaniti2018AutomatedLearning}, the deep-learning models for semantic segmentation usually require larger amounts of data. This fact has led to the segmentation of Gleason pattern $4$ and $5$ in a unified class, using low grade (Gleason Pattern $3$) or high grade (Gleason pattern $4$ or $5$). The main state-of-the-art convolutional neural networks (CNNs) have been used for this task. In particular, in \cite{Li2017AProstatectomies} a multi-resolution modification of the \textit{U-Net} architecture is proposed, while in \cite{Ing2018SemanticNetworks} the architectures proposed are the \textit{Fully-Convolutional Networks} and the \textit{SegNet}. Finally, a recent work proposed \textit{Region-CNNs} for both segmentation and glandular structures detection \cite{Li2019PathImages}.

In this work, we explore the automatic detection and grading of prostate tumour growth patterns by means of semantic segmentation of the Gleason grades in histology images. To the best of the authors' knowledge, this is the first time that automatic deep-learning segmentation models are used for the full gradation of cancerous patterns in prostate biopsies. One of the main contributions of this work is the validation of different well-known architectures for this task. In particular, we compare the performance of the \textit{Fully-Convolutional Networks}, the \textit{SegNet} and the \textit{U-Net} in the validation cohort. Furthermore, we propose the modification of the \textit{U-Net} architecture based on residual blocks for this task, outperforming previously mentioned architectures. This model has a comparable behaviour distinguishing between different Gleason grades than previous image-level approaches in the literature and offers an accurate delimitation of the different patterns.

\section{Materials}
\label{sec:materials}
The database used in this work is composed of $182$ prostate biopsies from $96$ patients. Whole Slides Images were obtained by staining and digitising the biopsies at $40\times$ magnification. The images were carefully analysed by a group of pathologists from Hospital Cl\'inico of Valencia, and pixel-level annotations were carried out following the Gleason grading system. In order to process the large Whole Slide Images, those were re-sampled to $10x$ resolution, and sliding-window patches of $512^2$ pixels and $50\%$ of overlap were obtained. For each image, a mask was extracted with the pixel-level semantic group among background (BG), non-cancerous tissue (NC), Gleason pattern $3$ (GP3), Gleason pattern $4$ (GP4) or Gleason pattern $5$ (GP5). Thus, the database is composed of $10339$ images with its respective semantic masks.   


\section{Methods}
\label{sec:methods}

The Gleason pattern grading of prostate images is addressed in this work by the pixel-level semantic segmentation using different convolutional-neural-networks models. Those are based on well-known architectures for image segmentation: \textit{Fully-Convolutional Networks}, \textit{SegNet}, and the \textit{U-Net}. The input images $x$ are resized during the training process to $256^2$ pixels in order to avoid memory problems. The proposed models in this work share the same output configuration: a convolutional layer with many filters as classes to be predicted. Concretely, the defined labels are: background (BG), non-cancerous tissue (NC), Gleason pattern $3$ (GP3), Gleason pattern $4$ (GP4) or Gleason pattern $5$ (GP5). Then, a pixel-level soft-max activation is used to obtain the probability maps. During the inference stage, a predicted map is obtained for each image assigning the class with a higher probability to each pixel.  

\subsection{Fully-convolutional networks}
\label{ssec:FCN}

\textit{Fully-Convolutional Networks} (FCN) were proposed in \cite{Long2015FullySegmentation} as an extension of classic classification architectures for semantic segmentation tasks. Convolutional neural networks (CNNs) for image classification are composed of a feature-extraction stage (base model) via stacked convolutional filters and spatial dimension reduction by max-pooling operations, and a classification phase through fully-connected layers (top model). In the FCN architecture, the top model is based on convolutional filters, providing a pixel-level prediction on the last activation maps. The main drawback of this architecture is the low resolution in the last activation map of the base model. For that purpose, pixel-level predictions at different pooling levels are combined. The lower pooling level used in the prediction is called the stride. The base model used for the feature extraction and the stride level define a concrete \textit{Fully-Convolutional Network} architecture (e.g. $FCN16$ for s stride of $16$).


\subsection{Segnet}
\label{ssec:segnet}

The \textit{SegNet} architecture \cite{Badrinarayanan2017SegNet:Segmentation} for semantic segmentation is based on the \textit{Fully-Convolutional Networks}. After the base model, a decoder branch recovers the spatial information via stacked convolutional blocks and upsampling operations. The upsampling is based on the indices used in the base model during the max-pooling operations in order to perform a non-linear reconstruction of the original dimensions.

\subsection{\textit{U-Net} architecture}
\label{ssec:unet}

\begin{figure}[htb]
    \begin{center}
    \includegraphics[width=.43\textwidth]{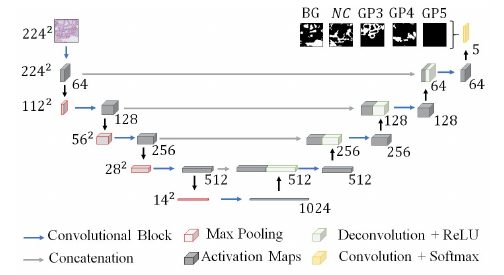}
    \caption{\textit{U-Net} architecture for prostate cancer gradation. BN: background, NC: non cancerous, GP3: Gleason pattern $3$, GP4: Gleason pattern $4$, GP5: Gleason pattern $5$.}
    \label{fig2}
    \end{center}
\end{figure}

The \textit{U-Net} architecture is a segmentation model proposed for medical applications in \cite{Ronneberger2015U-net:Segmentation}. The configuration is based on two branches: one encoder in charge of extracting the relevant features in the image, and a decoder controlling the reconstruction of the probability segmentation maps. The encoder branch consists of stacked convolutional blocks with dimensional reduction via a max-pooling operator. Each convolutional block doubles the number of filters, while the pooling operator resizes the image in a half. In particular, $4$ convolutional blocks are used, increasing the number of filters from $64$ up to $1024$ and the spatial dimensions from $256^2$ to $14^2$ pixels. In the decoder branch, the convolutional blocks are followed by deconvolutions that increase the dimension of the images in a factor of $2\times$ and reduce the number of filters in a half. Furthermore, the encoder is connected to the decoder via the concatenation of the activation maps of corresponding levels after the deconvolutional filter. An overview of the \textit{U-Net} used in this work is presented in Fig. \ref{fig2}. The convolutional block is composed of two convolutional filters of $3\times3$ pixels and ReLU as an activation function.

\subsection{\textit{U-Net} composed of residual blocks}
\label{ssec:residual}

\begin{figure}[htb]
    \begin{center}
    \includegraphics[width=.38\textwidth]{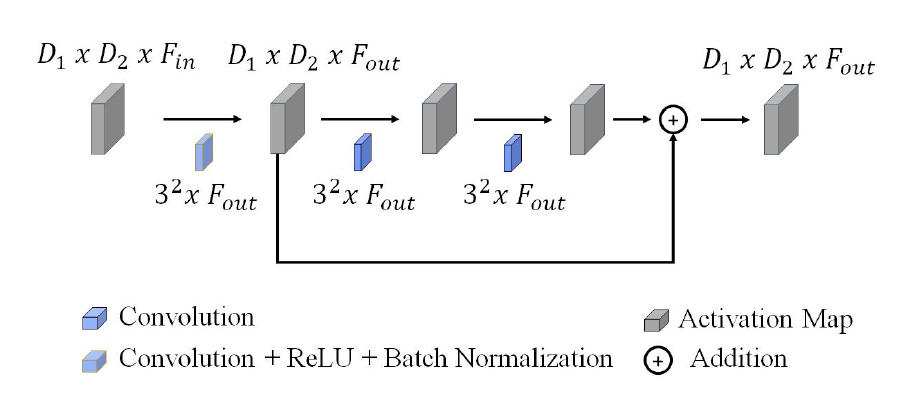}
    \caption{Residual Block with identity mapping modified for the \textit{U-Net} architecture. $F_{in}$: number of filters in the input volume. $F_{out}$: number of filters in the output volume. $D_{1}$, $D_{2}$: spatial dimensions of the activation volumes.}
    \label{fig3}
    \end{center}
\end{figure}

In order to improve the performance of the standard \textit{U-Net}, the convolutional blocks (see blue connections in Fig. \ref{fig2}) are modified with a residual configuration in the  $Res\textit{U-Net}$ architecture. The residual blocks \cite{He2016DeepRecognition} are a type of configuration of convolutional filters with skip-additive connections that have shown good properties for model optimisation and performance. In particular, the identity-mapping configuration proposed in \cite{He2016IdentityNetworks} is used. This is composed of three convolutional filters with a size of $3\times3$. The output of the first layer is connected in a skip connection with the result of processing a batch normalisation, ReLU activation and the other two filters to the same output. For the proposed \textit{U-Net} modification, a previous convolutional filter is used to normalise the number of filters (see Fig. \ref{fig3}).

\subsection{Loss function}
\label{ssec:loss}
 
The loss function used during the training process in the Dice function, introduced in \cite{vnet} for Volumetric Image Segmentation. This function makes a balance between intersection and union of predicted and reference masks, being appropriate for imbalanced datasets. The Dice is defined as follows:

\begin{equation}
Dice = \frac{2 \sum_{i}^{N} p_{i}g_{i}}{\sum_{i}^{N} p_{i}^{2} \sum_{i}^{N} g_{i}^{2}}
\end{equation}

\noindent where $p_{i}$ and $g_{i}$ denote the on-hot-encoded predicted labels and ground truth respectively for a batch of images for the class $i$. Note that $i$ denotes one of the $N$ classes: background, non cancerous, GP3, GP4, and GP5.

\section{Experiments and Results}
\label{sec:experimental}

In order to perform a validation and comparison of the different segmentation models described previously, the database was partitioned following a hold-out strategy. The images were divided into $3$ groups. Around the $60\%$ of the images were used for training, while two subsets with $20\%$ of the images were used for validation and testing. Note that the class balance was maintained among groups, and each patient was assigned uniquely to one group in order to avoid overestimation of the models' performance. As a figure of merit, the Dice index ($DI = 1 - Dice$) was obtained in the predicted segmentation maps.  

We trained $4$ types of convolution-neural-networks models for semantic segmentation of Gleason patterns. In particular, the \textit{Fully-Convolutional Network} ($FC8$) with pre-trained VGG16 weights as base model and stride of $8$, \textit{Segnet}, and \textit{U-Net} architecture with its standard configuration and the one modified with residual blocks $Res\textit{U-Net}$ were used. The hyperparameters were empirically optimised in the validation cohort. The $FCN8$ model was trained using an SGD optimiser with a Nesterov momentum of $0.9$, a learning rate of $1*10^{-2}$ and a decay rate of $1.6*10^{-3}$. In the \textit{U-Net} model the learning rate was fixed at $1*10^{-5}$, and Adam was used as the optimiser. Those models were trained during $200$ epochs in a mini-batch strategy of $16$ images. Regarding the \textit{Segnet} and $Res\textit{U-Net}$ models, Adam optimiser was also used, but the learning rate increased to $1*10^{-2}$. Those models were trained during $100$ epochs with a batch size of $8$ images. The results obtained in the validation subset are presented in Table \ref{tab1}.   

\begin{table}[htb]
\begin{center}
\caption{Results in the validation subset for the different models. BG: background, NC: non cancerous, GP: Gleason pattern.}
\label{tab1}
\resizebox{\linewidth}{!}{
\begin{tabular}{|l|l|l|l|l|l|}
\hline
\multicolumn{1}{|c|}{$\mathbf{Method}$} & \multicolumn{1}{c|}{$\mathbf{DI_{BG}}$}  & \multicolumn{1}{c|}{$\mathbf{DI_{NC}}$}  & \multicolumn{1}{c|}{$\mathbf{DI_{GP3}}$}  & \multicolumn{1}{c|}{$\mathbf{DI_{GP4}}$}  & \multicolumn{1}{c|}{$\mathbf{DI_{GP5}}$}    \\ \hline
$FCN8$ & $0.909$ & $0.087$ & $0.014$ & $0.660$ & $0.008$ \\ \hline
$SegNet$ & $0.966$ & $0.322$ & $0.706$ & $0.778$ & $0.520$ \\ \hline
$\textit{U-Net}$ & $0.975$ & $0.460$ & $0.468$ & $0.723$ & $0.265$ \\ \hline
$Res\textit{U-Net}$ & $\mathbf{0.972}$ & $\mathbf{0.464}$ & $\mathbf{0.796}$ & $\mathbf{0.820}$ & $\mathbf{0.535}$ \\ \hline
\end{tabular}
}
\end{center}
\end{table}

\begin{figure}[htb]
\captionsetup[subfloat]{farskip=1pt,captionskip=0.8pt}
    \centering
    
      \subfloat{\includegraphics[width=.28\linewidth, frame]{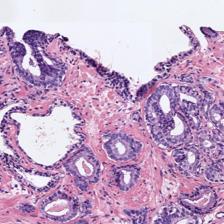}}
      \hspace*{\fill}
      \subfloat{\includegraphics[width=.28\linewidth, frame]{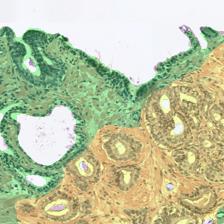}}
      \hspace*{\fill}
      \subfloat{\includegraphics[width=.28\linewidth, frame]{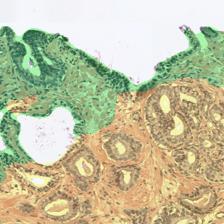}}
      \hspace*{\fill}    
    
      \subfloat{\includegraphics[width=.28\linewidth, frame]{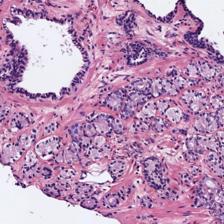}}
      \hspace*{\fill}
      \subfloat{\includegraphics[width=.28\linewidth, frame]{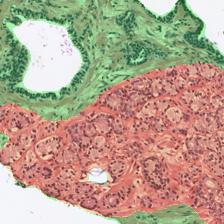}}
      \hspace*{\fill}
      \subfloat{\includegraphics[width=.28\linewidth, frame]{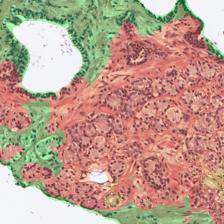}}
      \hspace*{\fill}
    
      \subfloat{\includegraphics[width=.28\linewidth, frame]{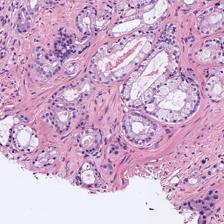}}
      \hspace*{\fill}
      \subfloat{\includegraphics[width=.28\linewidth, frame]{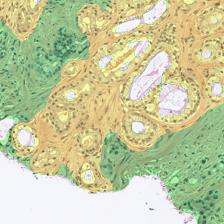}}
      \hspace*{\fill}
      \subfloat{\includegraphics[width=.28\linewidth, frame]{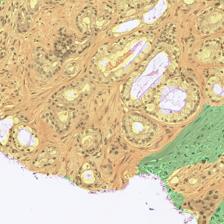}}
      \hspace*{\fill}
    
      \renewcommand{\thesubfigure}{a}
      \subfloat[\label{fig4a}]{\includegraphics[width=.28\linewidth, frame]{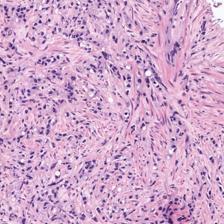}}
      \hspace*{\fill}
      \renewcommand{\thesubfigure}{b}
      \subfloat[\label{fig4b}]{\includegraphics[width=.28\linewidth, frame]{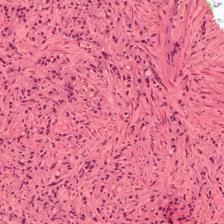}}
      \hspace*{\fill}
      \renewcommand{\thesubfigure}{c}
      \subfloat[\label{fig4c}]{\includegraphics[width=.28\linewidth, frame]{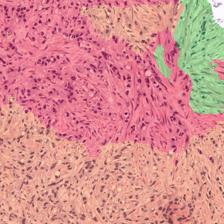}}
      \hspace*{\fill}
      
    \caption{Examples of our proposed $Res\textit{U-Net}$ performance
in the test set. Green: non cancerous, yellow: Gleason pattern $3$,  orange: Gleason pattern $4$ and red: Gleason pattern $5$. (a): Original Image, (b): Reference, (c): Predicted.}
    \label{fig4}
\end{figure}

Regarding the results obtained in the validation cohort, the \textit{U-Net} modified with residual blocks, $Res\textit{U-Net}$, showed the best performance. The worse performing model was the $FC8$, only able to recognise properly the tissue with Gleason pattern $4$. Better results were obtained with the \textit{Segnet} model than using the basic \textit{U-Net} architecture, with an average Dice index for the classes related to prostate tissue (i.e. NC, GP3, GP4 and GP5) of $0.5815$ and $0.4790$ respectively. The use of residual blocks showed to be crucial for the improvement of the \textit{U-Net} model, reaching an average Dice for these grades of $0.6538$.     
The best performing model, $Res\textit{U-Net}$, was trained in the whole training and validation set and the resultant model was evaluated in the test cohort. The obtained figures of merit and some representative examples of the semantic segmentation are presented in Table \ref{tab2} and Fig. \ref{fig4}, respectively. 

\begin{table}[htb]
\begin{center}
\caption{Results in the test set for the different models. BG: background, NC: non cancerous, GP: Gleason pattern.}
\label{tab2}
\resizebox{\linewidth}{!}{
\begin{tabular}{|l|l|l|l|l|l|}
\hline
\multicolumn{1}{|c|}{$\mathbf{Method}$} & \multicolumn{1}{c|}{$\mathbf{DI_{BG}}$}  & \multicolumn{1}{c|}{$\mathbf{DI_{NC}}$}  & \multicolumn{1}{c|}{$\mathbf{DI_{GP3}}$}  & \multicolumn{1}{c|}{$\mathbf{DI_{GP4}}$}  & \multicolumn{1}{c|}{$\mathbf{DI_{GP5}}$}    \\ \hline
$Res\textit{U-Net}$ & $0.982$ & $0.838$ & $0.419$ & $0.540$ & $0.194$\\ \hline
\end{tabular}
}
\end{center}
\end{table}

\begin{figure}[htb]
     
      \begin{center}
      \begin{tabular}{ c }
      \subfloat{\includegraphics[width=.70\linewidth]{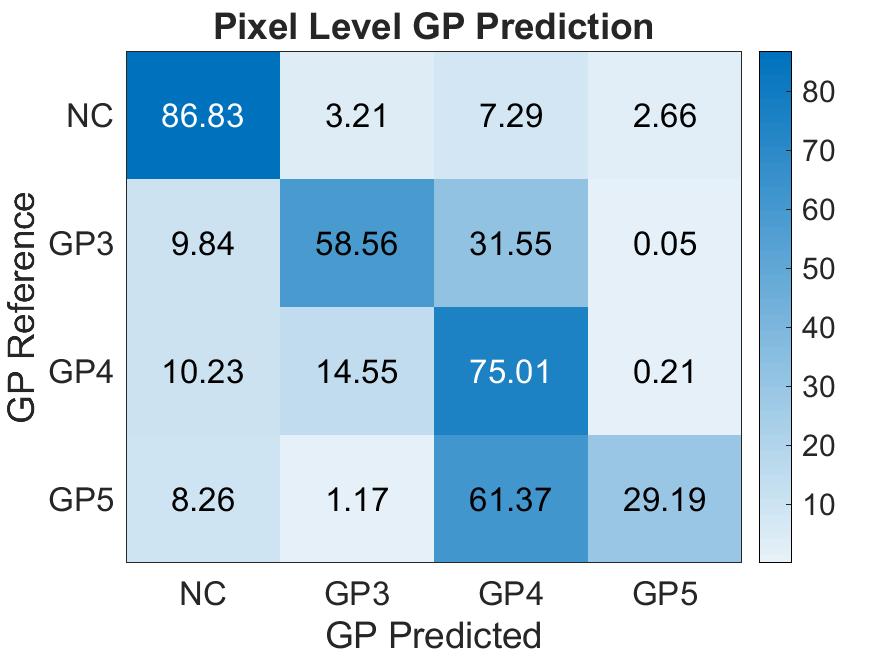}}
      \hspace*{\fill}
        \end{tabular}
      \end{center}
    \caption{Confusion matrix of the pixel-level Gleason grade prediction in the test cohort with the proposed $Res\textit{U-Net}$ model. NC: non cancerous, GP3: Gleason pattern $3$, GP4: Gleason pattern $4$, GP5: Gleason pattern $5$.}
    \label{fig5}
\end{figure}

The results obtained in the test subset show a slight decrease in model performance. The average Dice in the tissue classes drops to $0.4977$. This could be caused by the known internal heterogeneity in the Gleason grades, and the challenge of obtaining homogeneous subsets in the database during the partition stage. Moreover, the Dice index is a rigorous metric, and it does not take into account that most of the errors occur between adjacent classes (see Fig. \ref{fig4} example four). In previous literature related to image-level full Gleason gradation, the metric used is the quadratic Cohen's Kappa ($k$) \cite{Cohen1968WeightedCredit} to take into account this information. In order to establish fair comparisons with previous literature, the background class was joined to the non-cancerous class. The normalised confusion matrix is presented in Fig. \ref{fig5}, showing that most of the errors occur among adjacent classes and in pixels misclassified as cancerous due to a wrong delimitation of cancerous tissue (see Fig. \ref{fig4} examples one to three). The pixel-level $k$ value obtained was $0.52$, at the level of previous works in image-level approaches: $0.51$ in \cite{Arvaniti2018AutomatedLearning} in the test cohort or $0.61$ in \cite{Nir2019ComparisonImages} for the validation subset.         

\section{Conclusions}
\label{sec:conclusions}
In this research, we have proposed an \textit{U-Net} architecture modified with residual blocks able to perform semantic segmentation of the cancerous patterns in prostate images according to the Gleason grading system. The use of residual configurations is crucial to outperform other well-known architectures such that \textit{Segnet}. With the proposed model, a pixel-level Cohen's quadratic kappa of $0.52$ is reached in the test cohort. This performance is at the level of previous works for image-level grading of Gleason patterns, but our model offers a more accurate delimitation of cancerous patterns in the tissue. 

Further studies will focus on extensive comparison of the main three approaches for prostate image analysis for the full Gleason gradation: image-based, pixel-level segmentation methods and gland-level analysis.

\bibliographystyle{IEEEbib}
\bibliography{refs,references}

\begin{thebibliography}{10}

\bibitem{Ferlay2015Cancer2012}
Jacques Ferlay, Isabelle Soerjomataram, Rajesh Dikshit, Sultan Eser, Colin
  Mathers, Marise Rebelo, Donald~Maxwell Parkin, David Forman, and Freddie
  Bray,
\newblock ``{Cancer incidence and mortality worldwide: Sources, methods and
  major patterns in GLOBOCAN 2012},''
\newblock {\em International Journal of Cancer}, vol. 136, no. 5, pp.
  E359--E386, 2015.

\bibitem{Siegel2016Cancer2016}
Rebecca~L. Siegel, Kimberly~D. Miller, and Ahmedin Jemal,
\newblock ``{Cancer statistics, 2016},''
\newblock {\em CA: A Cancer Journal for Clinicians}, vol. 66, no. 1, pp. 7--30,
  2016.

\bibitem{gleason}
Donald Gleason,
\newblock ``Histologic grading of prostate cancer: A perspective, human
  pathology,'' 1992.

\bibitem{Litjens2016DeepDiagnosis}
Geert Litjens, Clara~I. S{\'{a}}nchez, Nadya Timofeeva, Meyke Hermsen, Iris
  Nagtegaal, Iringo Kovacs, Christina Hulsbergen-Van De~Kaa, Peter Bult, Bram
  Van~Ginneken, and Jeroen Van Der~Laak,
\newblock ``{Deep learning as a tool for increased accuracy and efficiency of
  histopathological diagnosis},''
\newblock {\em Scientific Reports}, vol. 6, no. January, pp. 1--11, 2016.

\bibitem{Arvaniti2018AutomatedLearning}
Eirini Arvaniti, Kim~S. Fricker, Michael Moret, Niels Rupp, Thomas Hermanns,
  Christian Fankhauser, Norbert Wey, Peter~J. Wild, Jan~H. R{\"{u}}schoff, and
  Manfred Claassen,
\newblock ``{Automated Gleason grading of prostate cancer tissue microarrays
  via deep learning},''
\newblock {\em Scientific Reports}, vol. 8, no. 1, pp. 1--11, 2018.

\bibitem{Nir2019ComparisonImages}
Guy Nir, Davood Karimi, S.~Larry Goldenberg, Ladan Fazli, Brian~F. Skinnider,
  Peyman Tavassoli, Dmitry Turbin, Carlos~F. Villamil, Gang Wang, Darby~J.S.
  Thompson, Peter~C. Black, and Septimiu~E. Salcudean,
\newblock ``{Comparison of Artificial Intelligence Techniques to Evaluate
  Performance of a Classifier for Automatic Grading of Prostate Cancer From
  Digitized Histopathologic Images},''
\newblock {\em JAMA network open}, vol. 2, no. 3, pp. e190442, 2019.

\bibitem{Li2019PathImages}
Wenyuan Li, Jiayun Li, Karthik~V. Sarma, King~Chung Ho, Shiwen Shen,
  Beatrice~S. Knudsen, Arkadiusz Gertych, and Corey~W. Arnold,
\newblock ``{Path R-CNN for Prostate Cancer Diagnosis and Gleason Grading of
  Histological Images},''
\newblock {\em IEEE Transactions on Medical Imaging}, vol. 38, no. 4, pp.
  945--954, 2019.

\bibitem{Ing2018SemanticNetworks}
Nathan Ing, Zhaoxuan Ma, Jiayun Li, Hootan Salemi, Corey Arnold, Beatrice~S.
  Knudsen, and Arkadiusz Gertych,
\newblock ``{Semantic segmentation for prostate cancer grading by convolutional
  neural networks},''
\newblock {\em Digital Pathology Conference}, vol. 1, no. June, pp. 46, 2018.

\bibitem{Li2017AProstatectomies}
Jiayun Li, Karthik~V Sarma, King~Chung Ho, Arkadiusz Gertych, Beatrice~S
  Knudsen, Corey~W Arnold, and Los Angeles,
\newblock ``{A Multi-scale U-Net for Semantic Segmentation of Histological
  Images from Radical Prostatectomies},''
\newblock {\em AMIA 2017 Annual Symposium}, , no. June 2019, pp. 1140--1148,
  2017.

\bibitem{Garcia2019First-stageLearning}
Gabriel Garc{\'{i}}a, Adrián Colomer, and Valery Naranjo,
\newblock ``{First-stage prostate cancer identification on histopathological
  images: Hand-driven versus automatic learning},''
\newblock {\em Entropy}, vol. 21, no. 4, 2019.

\bibitem{Long2015FullySegmentation}
Jonathan Long, Evan Shelhamer, and Trevor Darrell,
\newblock ``{Fully Convolutional Networks for Semantic Segmentation},''
\newblock {\em IEEE Conference on Computer Vision and Pattern Recognition
  (CVPR)}, vol. 1, pp. 1--10, 2015.

\bibitem{Badrinarayanan2017SegNet:Segmentation}
Vijay Badrinarayanan, Alex Kendall, and Roberto Cipolla,
\newblock ``{SegNet: A Deep Convolutional Encoder-Decoder Architecture for
  Image Segmentation},''
\newblock {\em IEEE Transactions on Pattern Analysis and Machine Intelligence},
  vol. 39, no. 12, pp. 2481--2495, 2017.

\bibitem{Ronneberger2015U-net:Segmentation}
Olaf Ronneberger, Philipp Fischer, and Thomas Brox,
\newblock ``{U-net: Convolutional networks for biomedical image
  segmentation},''
\newblock {\em Lecture Notes in Computer Science (including subseries Lecture
  Notes in Artificial Intelligence and Lecture Notes in Bioinformatics)}, vol.
  9351, pp. 234--241, 2015.

\bibitem{He2016DeepRecognition}
Kaiming He, Xiangyu Zhang, Shaoqing Ren, and Jian Sun,
\newblock ``{Deep residual learning for image recognition},''
\newblock {\em Proceedings of the IEEE Computer Society Conference on Computer
  Vision and Pattern Recognition}, vol. 2016-Decem, pp. 770--778, 2016.

\bibitem{He2016IdentityNetworks}
Kaiming He, Xiangyu Zhang, Shaoqing Ren, and Jian Sun,
\newblock ``{Identity mappings in deep residual networks},''
\newblock {\em Lecture Notes in Computer Science (including subseries Lecture
  Notes in Artificial Intelligence and Lecture Notes in Bioinformatics)}, vol.
  9908 LNCS, pp. 630--645, 2016.

\bibitem{vnet}
Fausto { Milletari1}, Nassir {Navab1}, and Seyed-Ahmad {Ahmadi},
\newblock ``V-net: Fully convolutional neural networks for volumetric medical
  image segmentation,''
\newblock 2016, vol.~39, pp. 2481--2495.

\bibitem{Cohen1968WeightedCredit}
Jacob Cohen,
\newblock ``{Weighted kappa: Nominal scale agreement provision for scaled
  disagreement or partial credit},''
\newblock {\em Psychological Bulletin}, vol. 70, no. 4, pp. 213--220, 1968.

\end{thebibliography}
\balance

\end{document}